\documentclass{article}
\usepackage{spconf,amsmath,graphicx}
\usepackage{xcolor}

\title{A novel repetition  normalized adversarial  reward for headline generation}
\name{Peng Xu {\normalfont and} Pascale Fung \thanks{Thanks to ITS/319/16FP of Innovation Technology Commission, HKUST 16248016 of Hong Kong Research Grants Council for funding.}}
\address{
  Department of Electronic and Computer Engineering\\
  Center for Artificial Intelligence Research (CAiRE)\\
  The Hong Kong University of Science and Technology, Clear Water Bay \\
  {\tt pxuab@ust.hk, pascale@ece.ust.hk }}
\begin{document}
%
\maketitle
\begin{abstract}
While reinforcement learning can effectively improve language generation models, it often suffers from generating incoherent and repetitive phrases \cite{paulus2017deep}. In this paper, we propose a novel repetition normalized adversarial reward to mitigate these problems. Our repetition penalized reward can greatly reduce the repetition rate and adversarial training mitigates generating incoherent phrases. Our model significantly outperforms the baseline model on ROUGE-1\,(+3.24),  ROUGE-L\,(+2.25), and a decreased repetition-rate (-4.98\%). 
\end{abstract}
\section{Introduction}
\label{sec:intro}

Summarization is the task of condensing a long document into a short summary without losing the main information. It has attracted lots of attention from the research community for its application to digest a large amount of information produced every day. Summarization can be generally categorized into two classes, extractive summary, and abstractive summary. Extractive summary \cite{nallapati2017summarunner} exclusively takes words from an input document and assembles them into a summary, while abstractive summary \cite{rush2015neural} needs to understand the document first and learns to paraphrase and generate new phrases. We focus on abstractive  summary in this paper.

Summarization models usually adopt maximum likelihood training. This training method suffers from two major disparities between training and testing, 1) exposure bias \cite{ranzato2015sequence}, that is during the training phases, the words from ground truth sentence are fed to the model while in the testing phase, the input to decoder comes from the prediction of generator. 2) different measurement between training and testing. In the training phase, the cost is measured by cross-entropy while in the testing phase, the model is evaluated with non-differentiable metrics, like ROUGE \cite{lin2004rouge} score. Reinforcement learning, as it can directly optimize non-differentiable metrics, has been proposed and successfully improved the performance of generation quality,  using ROUGE as reward \cite{paulus2017deep}. Nevertheless, reinforcement learning also suffers from generating incoherent and repeated phrases. Thus, a mixed training objective function and an intra-decoder is proposed in \cite{paulus2017deep}. From the perspective of reinforcement learning, we believe those two problems all come from a sub-optimal reward, ROUGE.  As ROUGE has no awareness of the quality of generated samples but just counts  n-gram matches, it enforces no penalty on repetition and incoherence.  For example, Table \ref{table: rouge_example} shows that a bad headline can achieve a high ROUGE score as it includes `` benson '' , `` for '', `` to '' and `` citigroup '' that overlap with real headlines. 


\begin{table}[t]
\begin{center}
\small
\resizebox{\linewidth}{!}{
\begin{tabular}{|l|}
\hline  \textbf{Article}: 
(...) He will miss the first \# minutes of the opening practice \\ session for the NAPA California \#  on Friday as a penalty for being \\ late to the pre-race inspection last Sunday at Alabama 's Talladega \\Superspeedway for the DieHard \# . It will mark the second \\
straight race in which Benson has missed practice time because \\ of a rules
infraction . (...) \\
\hline \textbf{RL:} benson to the for ford \\ 
\hline \textbf{Ground truth}: benson penalized for his bad timing \\
\hline \textbf{ROUGE-L} score: 0.358 \\ 

\hline  \textbf{Article}: (...) Vikram S. Pandit is doing some serious spring cleaning  \\ at Citigroup . Since becoming chief executive in December ,  Pandit \\ has been clearing out the corporate attic of weak businesses and \\ unloading worrisome assets  at bargain-basement prices .  (...) \\ 
\hline \textbf{RL:} sports column : citigroup to citigroup at citigroup \\ 
\hline \textbf{Ground truth}: citigroup embarks on plan to shed weak assets \\
\hline \textbf{ROUGE-L} score: 0.25 \\ \hline
\end{tabular}}
\end{center}
\caption{\label{table: rouge_example} Examples show that a bad headline with repetition and incoherence phrases can have high  ROUGE score.} 
\end{table}

In this paper, we propose a novel repetition normalized adversarial reward for reinforcement learning to mitigate the problems of incoherent and repeated headlines. We empirically show that our repetition penalized reward significantly reduces the repetition rate and adversarial training helps to generate more coherent phrases. 

\section{Related Work}
\label{sec:related_work}

Deep learning methods are first applied to two sentence-level abstractive summarization task on DUC-2004 and  Gigaword datasets \cite{rush2015neural} with an encoder-decoder model. This model is further extended by hierarchical network \cite{nallapati2016abstractive}, variational autoencoders \cite{miao2016language}, a coarse to fine approach \cite{tan2017neural} and minimum risk training \cite{shen2017recent}. As long summaries becomes more important, CNN/Daily Mail dataset was released in \cite{nallapati2016abstractive}. Pointer-generator with coverage loss \cite{see2017get} is proposed to approach the task by enabling the model to copy unknown words from article and penalizing the repetition with coverage mechanism. \cite{celikyilmaz2018deep} proposes deep communicating agents for representing a long document for abstractive summarization. There are more papers focusing on extractive summarizations\cite{nallapati2017summarunner, zhou2018neural}. Memory Network\cite{wu2018global, madotto2018mem2seq}, which can include external knowledge, might also be included into summarization model.

Reinforcement learning is also gaining popularity as it can directly optimize non-differentiable metrics \cite{pasunuru2018multi, paulus2017deep}. \cite{paulus2017deep}  proposes an intra-decoder model and combines reinforcement learning and maximum likelihood training to deal with summaries with bad qualities. We instead propose to improve headlines with a novel reward. Reinforcement learning is also explored with generative adversarial network (GAN) \cite{yu2017seqgan}. \cite{liu2017generative} applies the generative adversarial network for summarization to achieve a better performance. However, they directly take the score from  the discriminator as the reward while we combine it with repetition penalized ROUGE score. 

\section{Methodology}
\label{sec:methodology}
\begin{figure}[htb]

  \centering
  \centerline{\includegraphics[width=8.5cm]{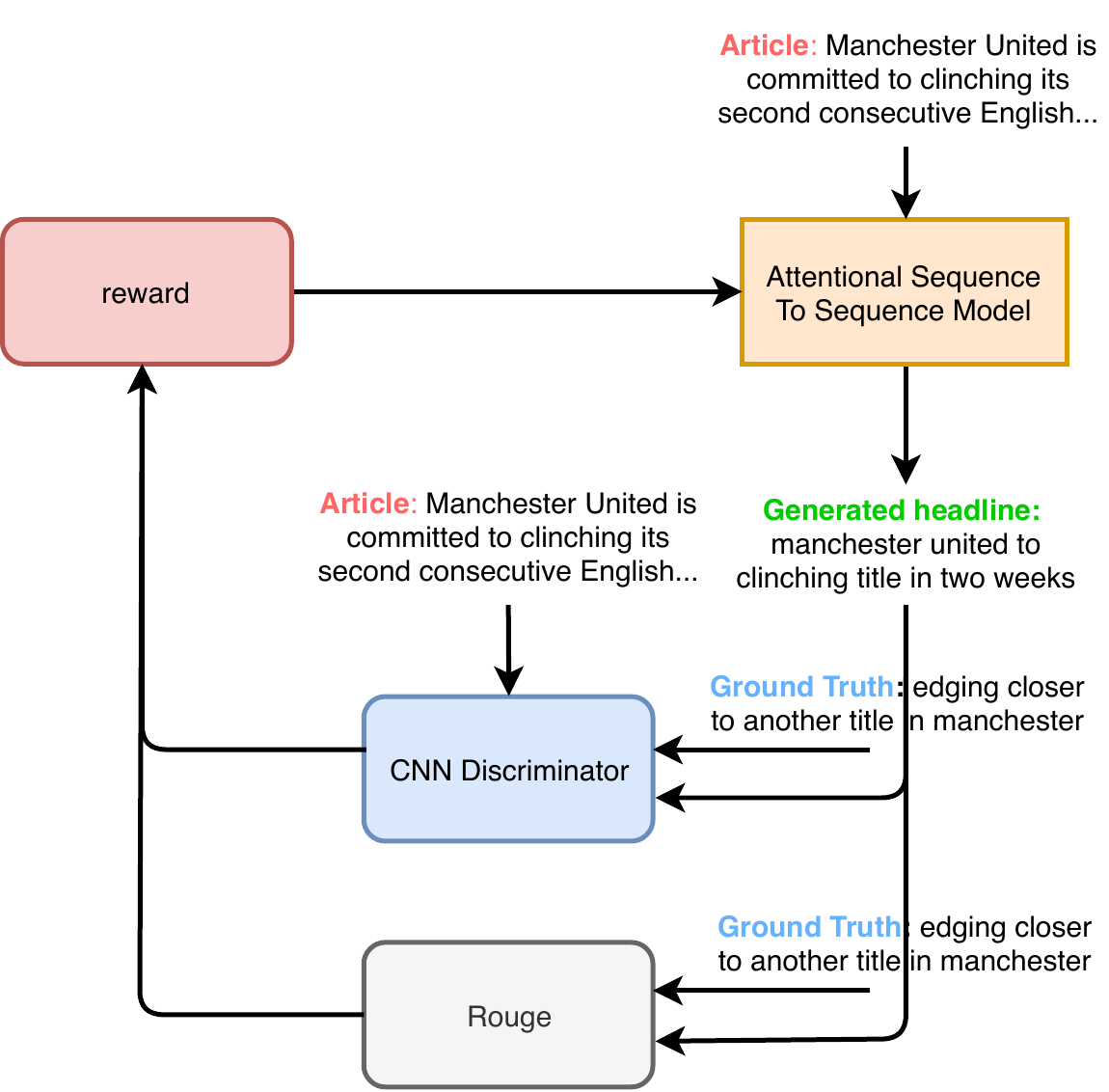}}
  \centerline{}\medskip
\caption{Diagram of our learning framework}
\label{fig:adv_normalized}
\end{figure}

Our model consists of two parts, an attentional sequence to sequence model with pointer-generator and a Convolutional Neural Network (CNN) discriminator. The attentional sequence to sequence model takes a news article as input and generates a headline. The discriminator takes either the generated headline or real headline as input and outputs a probability of how likely  the generated headline is real. This probability is then combined with the ROUGE score between the generated sample and the real one, as the reward for our sequence to sequence model. The model is shown in Figure \ref{fig:adv_normalized}. 

\subsection{Attentional sequence to sequence model with pointer-generator}
We abbreviate this attentional sequence to sequence model with pointer-generator \cite{see2017get} as Pointer-Gen-Coverage. The model is described below. Firstly, the tokens of each article $w_i$ are fed into the encoder one-by-one and the encoder generates a sequence of hidden states $ h_i $. For each decoding step $t$, the decoder receives the embedding for each token of a headline as input, and updates its hidden states $s_t$. The attention mechanism is calculated as in \cite{bahdanau2014neural} \:
\begin{align}
    e_i^t &= v^T \text{tanh}(W_h h_i + W_s s_t + w_c c_i^t + b_{attn}) \\
    a^t &= \text{softmax}(e^t) \\
    h_t^* &= \sum\nolimits_i a_i^t h_i 
\end{align}
where $W_h$, $W_s$, $w_c$, $b_{attn}$, $v$ are the trainable parameters, $h_t^*$ is the context vector, $c_i^t$ is the coverage vector defined below.  $s_t$, $h_t^*$ are then combined to predict the next word. 
\begin{align}
    c_i^t & = \sum\nolimits_{t^{'}=0}^{t-1} a_i^{t^{'}} \\
    o_t & = V ([s_t, h_t^*]) + b \\
    P_{vocab} & = \text{softmax} ( V^{'} o_t + b^{'})
\end{align}
where $V$, $b$, $V^{'}$, $b^{'}$ are parameters to train.  $c_i^t$ is defined as the accumulated attention over specific positions. We also include pointer generator network to enable our model to copy rare/unknown words from input article.

\begin{align}
    p_{gen} & = \sigma (w_{h^*}^T h_t^* + w_s^T s_t + w_x^T x^t + b_{ptr}) \\
    P(w) & = p_{gen} P_{vocab}(w) + (1 - p_{gen}) \sum\nolimits_{i:w_i=w} a_i^t
\end{align}
where $x^t$ is the embedding of input word of decoder, $w_{h^*}^T$, $w_s^T$, $w_x^T$, $b_{ptr}$ are trainable parameters. Our final loss function for Maximum Likelihood (ML) training  thus becomes: 
\begin{align}
    L_{\text{ml}} = - \frac{1}{T} \sum\nolimits_{t=1}^T (\text{log} P(w_t) + \lambda \sum\nolimits_i \text{min}(a_i^t, c_i^t))
\end{align}

\subsection{Adversarial training}
To measure the quality of generated sample, we train a CNN discriminator to classify whether the sample is generated or a real one. Our CNN model takes an article $A$ and real headline or generated headline $H$ as input. We employ two CNNs with same structures \cite{kim2014convolutional} to encode $H$ and $A$ respectively. The features are then concatenated and then projected to one single scalar $D(A, H)$ with a sigmoid activation, as the score of the headline being a real one. The loss function is constructed to maximize the log likelihood of real samples and minimize that of generated samples.
\begin{align}
L_{d} & = - E_{H \sim p_{data}} \text{log} D(A, H) \\  
&\phantom{{}=-}- E_{H \sim P(w)} \text{log} (1 - D(A, H)) \notag
\end{align}
\subsection{Repetition normalized adversarial reward}
\label{sec:reward}
Our Repetition normalized adversarial reward (ROUGE-RP-ADV) is the harmonic mean of repetition penalized ROUGE score (ROUGE-RP) and CNN discriminator  score $ D(A, H)$ by further introducing $\beta$ to balance ROUGE-RP and $D(A, H)$. Larger $\beta$ encourages our model to emphasize  ROUGE-RP.  Repetition-rate, ROUGE-RP and  ROUGE-RP-ADV are defined below. 
\begin{align}
\text{repetition-rate} &= 1. -  \frac{N(\text{unique tokens of H})}{N(\text{total tokens of H})} \\
\text{ROUGE-RP} &=  (1 - \text{repetition-rate}) * \text{ROUGE} \\
\text{ROUGE-RP-ADV} &= \frac{(1 + \beta^2) \, \text{ROUGE-RP} * D(A, H)} {\text{ROUGE-RP} + \beta^2  D(A, H) }
\end{align}
where $N$ counts the number of tokens. 

\subsection{Reinforcement Learning}
We use REINFORCE algorithm \cite{williams1992simple} and baseline model proposed in \cite{ranzato2015sequence} to train our generator (Pointer-Gen-Coverage). In each training step, a sentence is first sampled based on the $P(w)$ from our generator. A reward $R$ is then calculated between generated sample and real headline. For each time $t$, a linear regression model is utilized to estimate the reward of step $t$ based on $t$-th  state $o_t$. The linear regression model and loss function is shown below:
\begin{align}
\hat{R_t} &= W_r  o_t + b_r \\
L_b &= \frac{1}{T} \sum\nolimits_{t=1}^T||  R - \hat{R_t} ||^2
\end{align}
where $W_r$ and $b_r$ are trainable parameters, $R$ is the reward for whole sentence. Our final loss function for reinforcement learning (RL) becomes:
\begin{align}
    L_{\text{RL}} = - \frac{1}{T} \sum\nolimits_{t=1}^T  (R - \hat{R_t}) \, \text{log} P(w_t) 
\end{align}

\begin{table*}[ht]
\centering
\resizebox{0.9\textwidth}{!}{
\begin{tabular}{|c|cccc|}
\hline
\bf models                              & \bf ROUGE-1    &  \bf ROUGE-2    & \bf ROUGE-L     & \bf repetition-rate \\ \hline
Pointer-Gen-Coverage                            & 24.68   & 10.92 & 21.78 & 9.54 \%                                      \\
Pointer-Gen-Coverage + ROUGE             & \bf{28.65}    & 9.70 & 23.89  & 13.42\%                                    \\
Pointer-Gen-Coverage + ROUGE + MLE                      & 26.64 & \bf{11.87}  & 23.54 & 10.28\%  \\
Pointer-Gen-Coverage + ROUGE-RP     & 27.51     &  9.39  & 23.54  & \bf{4.23\%}                                                \\
Pointer-Gen-Coverage + ROUGE-RP-ADV  & 27.92  &  10.27      &  \bf{24.03}   & 4.56\%                                              \\ \hline

\end{tabular}
}
\caption{Results of different models on f-score for different ROUGE measures and repetition rate. The larger ROUGE score  and smaller repetition rate implies better results. }
\label{table:result}
\end{table*}
\section{Experiments}
\label{sec:experiments}

\subsection{Dataset}
\label{sec:dataset}
Recent neural headline generation models focus on generating headlines from selected recapitulative sentences. However, these selected sentences may not have enough information for the generation, as for example, in New York Times, the overlap between headlines and the sentences is very low  \cite{tan2017neural}. Thus, in this paper, we focus on generating headlines from full document. We use the dataset of New York Times part in Gigaword \cite{napoles2012annotated}. Different from \cite{rush2015neural, nallapati2016abstractive}, we use the whole document as our input. We follow the preprocessing steps \footnote{https://github.com/facebookarchive/NAMAS} in \cite{rush2015neural} and we then use the NLTK \cite{bird2004nltk} to tokenize our dataset. The average length of headlines and articles are 863.4 and 8.6, respectively.  We empirically choose 400 words as the maximum article length as it covers 67\% tokens of headlines while full documents achieve 73\% overlap. Following \cite{tan2017neural}, we randomly split our train, dev, and test set as 1106824, 2000 and 2000.

\subsection{Training Details}
\label{sec:dataset}
\noindent\textbf{ML learning:} Our reinforcement learning model is first pretrained by optimizing $L_{\text{ml}}$. Adam optimizer is used and learning rate is 0.0001. Batch size is set as 16 and one layer, bi-directional Long Short-Term Memory (bi-LSTM) \cite{hochreiter1997long} model with 512 hidden sizes and 100 embedding size is utilized. Gradients with l2 norm larger than 2.0 are clipped.  We stop training when the ROUGE-1 f-score stops increasing.  Beam-search with the beam size of 5 is adopted for decoding.

\noindent\textbf{Adversarial Training:} As our RL model is well pretrained, CNN discriminator also needs to be pretrained to avoid an imbalanced generator and discriminator. CNN discriminator is trained by optimizing $L_d$ by one epoch. We use one layer CNN model with filter sizes of 1, 3, and 5. Each channel contains 512 filters. Adam optimizer with the learning rate of 0.001 is used. After training, our CNN discriminator achieves the accuracy of 0.6945 on real headlines and accuracy of 0.6975 on generated headlines. 

\noindent\textbf{RL training:} We found that using reward of ROUGE-1 f-score will always reach the maximum decoding length, thus the f-score for ROUGE-L is utilized as the ROUGE reward. For Pointer-Gen-Coverage model, we use the Adam optimizer with a learning rate of 0.0001. When ROUGE-RP-ADV is used as the reward,  our best model is achieved when $\beta$ is 2000. When mixing ML with RL \cite{paulus2017deep}, a large weight $\alpha$ for RL is necessary to achieve good performance and the best model is acquired with $\alpha$ set as 0.97.

\begin{table}[ht]
\small
\resizebox{\linewidth}{!}{
\begin{tabular}{|l|} \hline
Pointer-Gen-Coverage (\bf 26.43): \\ \textbf{manchester united to retain second english premier title}  \\ \hline
Pointer-Gen-Coverage+ROUGE (\bf 52.86): \\ \textbf{ \textcolor{red}{manchester united} to second title in \textcolor{red} {manchester united}} \\ \hline
Pointer-Gen-Coverage+ROUGE-RP (\bf 28.57) : \\ \textbf{manchester united to second english premier title} \\ \hline
Pointer-Gen-Coverage+ROUGE-RP-ADV (\bf 39.65) : \\ \textbf{manchester united to clinching title in two weeks} \\ \hline
Pointer-Gen-Coverage+ROUGE+MLE (\bf 26.43) :  \\ \textbf{manchester united to keep \# nd english title}\\ \hline
Ground Truth : \\\textbf{ edging closer to another title in manchester } \\ \hline
\end{tabular}}

\caption{\label{table: example}  An example of headlines generated by different models. The ROUGE-L score is reported inside parenthesis. Pointer-Gen-Coverage+ROUGE+MLE generates a headline with 26.43 ROUGE-L score. Pointer-Gen-Coverage+ROUGE gives 52.86 ROUGE-L score with repetition of ``manchester united'', Pointer-Gen-Coverage+ROUGE-RP gives incoherent phrases with a score of 28.57, and Pointer-Gen-Coverage+ROUGE-RP-ADV generates a more coherent headline and achieves the high ROUGE-L score of 39.65.}
\end{table}
\subsection{Results}
\label{sec:result}

We report f-score for ROUGE-1, ROUGE-2, ROUGE-L on the test set. Table \ref{table:result} shows the results of different models.  The baseline Pointer-Gen-Coverage model achieves 
24.68  ROUGE-1 score, 10.92 ROUGE-2 score, 21.78 ROUGE-L score, and repetition-rate of 9.54\%. When applying ROUGE score alone as the reward, ROUGE-1 score increases to 28.65, and ROUGE-L increases to 23.89. However, the repetition-rate also increases to 13.42\%, which shows that using ROUGE alone as reward improves the performance but also introduces more repetitions. 

When using repetition penalized rouge reward, repetition rate decreases from 13.42\% to 4.23\%. It implies that by adding the penalty to reward, our model learns to generate headlines with fewer repetitions. However, we observe that both Pointer-Gen-Coverage+ROUGE and Pointer-Gen-Coverage+ROUGE-RP model produce incoherent headlines like `` for opera singer , a tenor to the '', which ends the headline abruptly. By combining ROUGE score with CNN discriminator score, our Pointer-Gen-Coverage+ROUGE-RP-ADV model generates more natural headlines and achieves ROUGE-1 27.92,  ROUGE-L 24.03, and a decreased repetition-rate of 4.56\%. It outperforms the baseline model of Pointer-Gen-Coverage on ROUGE-1\,(+3.24),  ROUGE-L\,(+2.25), and a decreased repetition-rate (-4.98\%).  An example is shown in  Table \ref{table: example} to demonstrate the differences. We also compare our Pointer-Gen-Coverage+ROUGE-RP-ADV model to the model with a mixed training objective function (Pointer-Gen-Coverage+ROUGE+
MLE), which is introduced in \cite{paulus2017deep} to deal with incoherent generations. Our model achieves better results on ROUGE-1\,(+1.28), ROUGE-L\,(+0.49), and repetition-rate (-5.72\%). 

To further understand how the repetition penalized reward reduces generations of repeated phrase, we calculate the ROUGE-RP score for Pointer-Gen-Coverage+ROUGE  and Pointer-Gen-Coverage+ROUGE-RP respectively and we get 20.91 and 22.58.   Compared to Pointer-Gen-Coverage+ROUGE, Pointer-Gen-Coverage+ROUGE-RP achieves lower ROUGE-L score (23.54  vs 23.89) but higher ROUGE-RP score (22.58 vs 20.91) and lower repetition rate. This illustrates that, our model is encouraged to sacrifice ROUGE-L score for repetition avoidance. For the adversarial training, we take the trained CNN discriminator out and feed it with the model outputs of Pointer-Gen-Coverage+ROUGE-RP and Pointer-Gen-Coverage+ROUGE-RP-ADV in Table \ref{table: example}. The scores for Pointer-Gen-Coverage+ROUGE-RP, Pointer-Gen-Coverage+ROUGE-RP-ADV and ground truth are 0.9993, 0.9996, and 1.0. Our CNN discriminator believes Pointer-Gen-Coverage+ROUGE-RP-ADV generates a better headline than Pointer-Gen-Coverage+ROUGE-RP.

\section{Conclusion}
\label{sec:conclusion}
In this paper, we proposed a repetition normalized adversarial reward for reinforcement learning on headline generation. We empirically showed that our repetition penalized reward greatly decreased the repetition rate of the generated headlines and adversarial training further helped the model generate more natural headlines. Experiments showed that our model outperformed the baseline model on ROUGE-1\,(+3.24),  ROUGE-L\,(+2.25), and a decreased repetition-rate (-4.98\%).



\bibliographystyle{IEEEbib}
\bibliography{strings,refs}

\end{document}